# Recoverability of Joint Distribution from Missing Data

**Jin Tian**
Department of Computer Science
Iowa State University
Ames, IA 50014
`jtian@iastate.edu`

## Abstract

A probabilistic query may not be estimable from observed data corrupted by missing values if the data are not missing at random (MAR). It is therefore of theoretical interest and practical importance to determine in principle whether a probabilistic query is estimable from missing data or not when the data are not MAR. We present an algorithm that systematically determines whether the joint probability is estimable from observed data with missing values, assuming that the data-generation model is represented as a Bayesian network containing unobserved latent variables that not only encodes the dependencies among the variables but also explicitly portrays the mechanisms responsible for the missingness process. The result significantly advances the existing work.

## 1 Introduction

Missing data occur when some variable values are missing from recorded observations. It is a common problem across many disciplines including artificial intelligence, machine learning, statistics, economics, and the health and social sciences. Missing data pose a major obstacle to valid statistical and causal inferences in a broad range of applications. There is a vast literature on dealing with missing data in diverse fields. We refer to [1, 2] for a review of related work. Most work in machine learning assumes data are *missing at random (MAR)* [3, 4], under which likelihood-based inference (as well as Bayesian inference) can be carried out while ignoring the mechanism that leads to missing data.

In principle, however, to analyze data with missing values, we need to understand the mechanisms that lead to missing data, in particular whether the fact that variables are missing is related to the underlying values of the variables in the data set. Indeed some work in machine learning explicitly incorporates missing data mechanism into the model [5, 6, 7, 8]. Recently Mohan et al. [1] have used directed acyclic graphs (DAGs) to encode the missing data model, called *m-graphs*, by representing both conditional independence relations among variables and the mechanims responsible for the missingness process. M-graphs provide a general framework for inference with missing data when the MAR assumption does not hold and the data are categorized as *missing not at random (MNAR)*. Whether a hypothesized DAG model or MAR assumption is testable with missing data is investigated in [9, 10]. A graphical version of MAR defined in terms of graphical structures is discussed in [10].

One important research question under this graphical model framework is: Is a target probabilistic query estimable from observed data corrupted by missing values given a missing data model represented as a m-graph? It is known that when the data are MAR, the joint distribution is estimable. On the other hand, when the data are MNAR, a probabilistic query may or may not be estimable depending on the query and the exact missing data mechanisms. For example, consider a single random variable $X$ and assume that whether the values of $X$ are missing is related to the values of $X$ (e.g., in a salary survey, people with low income are less likely to reveal their income). The model is

MNAR and we can not estimate $P(X)$ unbiasedly even if infinite amount of data are collected. In practice it is important to determine in principle whether a target query is estimable from missing data or not. Several sufficient graphical conditions have been derived under which probability queries of the form $P(x, y)$ or $P(y|x)$ are estimable [1]. Mohan and Pearl [2] extended those results and further developed conditions for recovering causal queries of the form $P(y|do(x))$. Shpitser et al. [11] formulated the problem as a causal inference problem and developed a systematic algorithm for estimating the joint distribution when the model contains no unobserved latent variables.

In this paper we develop an algorithm for systematically determining the recoverability of the joint distribution from missing data in m-graphs that could contain latent variables. The result is significantly more general than the sufficient conditions in [1, 2]. Compared to the result in [11] we allow latent variables in the model, and treat the problem in a purely probabilistic framework without appealing to causality theory.

The paper is organized as follows. Section 2 defines the notion of m-graphs as introduced in [1]. Section 3 formally defines the notion of recoverability and briefly reviews previous work. In Section 4 we present our algorithm for recovering the joint distribution. Section 5 concludes the paper.

## 2  Missing data model as a Bayesian network

Bayesian networks are widely used for representing data generation models [12, 13]. Mohan et al. [1] used DAGs called *m-graphs*, to represent both the data generation model and the mechanisms responsible for the missingness process. In this section we define m-graphs, mostly following the notation in [1].

Let $G$ be a DAG over a set of variables $V \cup L \cup R$ where $V$ is the set of observable variables, $L$ is the set of unobserved latent variables, and $R$ is the set of missingness indicator variables introduced in order to represent the mechanisms that are responsible for missingness. We assume that $V$ is partitioned into $V_o$ and $V_m$ such that $V_o$ is the set of variables that are observed in all data cases and $V_m$ is the set of variables that are missing in some data cases and observed in other cases.[1] Every variable $V_i \in V_m$ is associated with a variable $R_{V_i} \in R$ such that, in any observed data case, $R_{V_i} = 1$ if the value of corresponding $V_i$ is missing and $R_{V_i} = 0$ if $V_i$ is observed. We require that $R$ variables may not be parents of variables in $V$, since $R$ variables are missingness indicator variables and we assume that the data generation process over $V$ variables does not depend on the missingness mechanism. For any set $S \subseteq V_m$, let $R_S$ represent the set of $R$ variables corresponding to variables in $S$.

The DAG $G$ provides a compact representation of the missing data model $P(V, L, R) = P(V, L)P(R|V, L)$, and will be called a m-graph of the model. The m-graph depicts both the dependency relationships among variables in $V \cup L$ and the missingness mechanisms, and it encodes conditional independence relationships that can be read off the graph by d-separation criterion [14] such that every d-separation in the graph $G$ implies conditional independence in the distribution $P$. See Figure 1 for examples of m-graphs, in which we use solid circles to represent always observed variables in $V_o$ and $R$, and hollow circles to represent partially observed variables in $V_m$. We often use bidirected edges as a shorthand notation to denote the existence of a $L$ variable as common parent of two other variables. See Figures 1(b) for an example.

## 3  Recoverability

Given a m-graph and observed data with missing values, it is important to know whether we can in principle compute a consistent estimate of a given probabilistic query $q$ (e.g. $P(x|y)$). If $q$ is deemed to be not estimable (or recoverable) then it is not estimable even if we have collected infinite amount of data. Next we formally define the notion of recoverability.

In any observation, let $S \subseteq V_m$ be the set of observed variables (i.e., values of variables in $V_m \setminus S$ are missing). Then the observed data is governed by the distribution $P(V_o, S, R_{V_m \setminus S} = 1, R_S = 0)$. Formally

---

[1]We assume we could partition the $V$ variables into $V_o$ and $V_m$ based on domain knowledge (or modeling assumption). In many applications, we have the knowledge that some variables are always observed in all data cases.

**Definition 1 (Recoverability)** *Given $m$-graph $G$, a target probabilistic query $q$ is said to be recoverable if $q$ can be expressed in terms of the set of observed positive probabilities $\{P(V_o, S, R_{V_m \setminus S} = 1, R_S = 0) : S \subseteq V_m\}$ - that is, if $q^{M_1} = q^{M_2}$, for every pair of models $P^{M_1}(V, R)$ and $P^{M_2}(V, R)$ compatible with $G$ with $P^{M_1}(V_o, S, R_{V_m \setminus S} = 1, R_S = 0) = P^{M_2}(V_o, S, R_{V_m \setminus S} = 1, R_S = 0) > 0$ for all $S \subseteq V_m$.*

This collection of observed probabilities $\{P(V_o, S, R_{V_m \setminus S} = 1, R_S = 0) : S \subseteq V_m\}$ has been called the *manifest distribution* and the problem of recovering probabilistic queries from the manifest distribution has been studied in [1, 9, 2, 11].

**Example 1** *In Fig. 1(a), the manifest distribution is the collection $\{P(X, Y, R_X = 0, R_Y = 0), P(X, R_X = 0, R_Y = 1), P(Y, R_X = 1, R_Y = 0), P(R_X = 1, R_Y = 1)\}$.*

### 3.1 Previous work

When data are MAR, it is known that the joint distribution is recoverable. We have $R \perp\!\!\!\perp V_m | V_o$[2] (see [1, 10] for graphical definition of MAR), and the joint is recoverable as $P(V) = P(V_m|V_o)P(V_o) = P(V_m|V_o, R = 0)P(V_o)$.

When data are MNAR, the joint $P(V)$ may or may not be recoverable depending on the m-graph $G$. Mohan and Pearl [2] presented a sufficient condition for recovering probabilistic queries including joint by using sequential factorizations (extending ordered factorizations in [1]). The basic idea is to find an order of variables, called *admissible sequence*, in $V \cup R$ such that $P(V)$ could be decomposed into an ordered factorization or sum of it such that every factor is recoverable by using conditional independence relationships.

**Example 2** *We want to recover $P(X, Y)$ given the m-graph in Fig. 1(a). The order $X < R_Y < Y$ induces the following sum of factorization:*

$$P(x, y) = \sum_{r_Y} P(x|r_Y, y)P(r_Y|y)P(y) = P(y) \sum_{r_Y} P(x|r_Y)P(r_Y), \qquad (1)$$

*where both $P(y) = P(y|R_Y = 0)$ and $P(x|r_Y) = P(x|r_Y, R_X = 0)$ are recoverable.*

The main issue with the sequential factorization approach is that it is not clear in general whether an admissible sequence exists or how to find an admissible sequence (even deciding whether a given order is admissible or not does not appear to be easy). Several sufficient conditions for recovering the joint $P(V)$ are given in [1, 2] which may handle problems for which no admissible sequence exists. For example, one condition says $P(V)$ is recoverable if no variable $X$ is a parent of its corresponding $R_X$ and there are no edges between the $R$ variables.

If the m-graph does not contain latent variables ($L = \emptyset$), the model is called a *Markov model*. In a Markov model, applying the chain rule of Bayesian network, we have

$$P(v_o, v_m, R = 0) = P(v)P(R = 0|v) = \prod_i P(v_i|pa_i) \prod_j P(R_{V_j} = 0|pa_{r_{V_j}}) \big|_{R=0} \qquad (2)$$

where $Pa_i$ and $Pa_{r_{V_j}}$ represent the parents of $V_i$ and $R_{V_j}$ in $G$ respectively. From Eq. (2) we obtain that $P(V) = \prod_i P(v_i|pa_i)$ is recoverable if every factor $P(R_{V_j} = 0|pa_{r_{V_j}})$ is recoverable. Shpitser et al. [11] developed a systematic algorithm MID for recovering $P(V)$ by trying to recover every $P(R_{V_j} = 0|pa_{r_{V_j}})$ using a subroutine DIR. The MID/DIR algorithm was based on formulating the recoverability problem as a causal inference problem and using techniques developed for the problem of identification of causal effects.

Allowing latent variables in the model, however, makes the recoverability problem substantially different (more difficult).

---

[2]We use $X \perp\!\!\!\perp Y | Z$ to denote that $X$ is conditionally independent of $Y$ given $Z$.

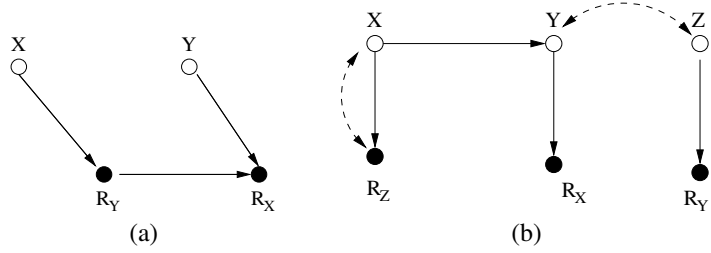

Figure 1: (a) A m-graph that is MNAR. $P(X,Y)$ is recoverable. (b) A m-graph containing latent variables. We use solid circles to represent always observed variables in $V_o$ and $R$, and hollow circles to represent partially observed variables in $V_m$. We use bidirected edges to denote the existence of a latent $L$ variable as common parent of two variables.

**Example 3** *In Fig. 1(b), applying the chain rule of Bayesian network, we have*

$$P(x,y,z,R_{X,Y,Z}=0)$$
$$=[\sum_{l_1} P(y|x,l_1)P(z|l_1)P(l_1)][\sum_{l_2} P(x|l_2)P(R_Z=0|l_2)P(l_2)]P(R_X=0|y)P(R_Y=0|z), \quad (3)$$

*while*

$$P(x,y,z) = P(x)[\sum_{l_1} P(y|x,l_1)P(z|l_1)P(l_1)]. \quad (4)$$

*It is clear that we cannot recover $P(x,y,z)$ by trying to recover every $P(R_{V_j}=0|pa_{r_{V_j}})$, as $P(v)$ and $P(R=0|v)$ do not decouple as in Eq. (2) anymore. MID algorithm is not applicable anymore.*

In this paper we deal with the problem of recovering the joint $P(V)$ in models that may contain latent variables. We will treat the problem in a purely probabilistic framework without appealing to causality theory.

## 4 Recoverability of the joint

In this section we develop an algorithm that will systematically determine the recoverability of the joint $P(V)$. First we reformulate the given observed probabilities.

**Proposition 1** *Given the manifest distribution $\{P(V_o, S, R_{V_m \setminus S} = 1, R_S = 0) : S \subseteq V_m\}$, the probability $P(V_o, S, R_{V_m \setminus S}, R_S = 0)$ is recoverable for all $S \subseteq V_m$.*

*Proof:* For any $r_{V_m \setminus S}$ values, let $T$ be the set of variables in $V_m \setminus S$ for which $r_T = 0$, then $r_{V_m \setminus (S \cup T)} = 1$. We have $P(V_o, S, R_{V_m \setminus S}, R_S = 0)$ is recoverable as

$$P(v_o, s, r_{V_m \setminus S}, R_S = 0) = \sum_t P(v_o, s \cup t, R_{V_m \setminus (S \cup T)} = 1, R_{S \cup T} = 0) \quad (5)$$

□

It turns out that it is much easier to work with the set of probabilities $P(V_o, S, R_{V_m \setminus S}, R_S = 0)$ than with $P(V_o, S, R_{V_m \setminus S} = 1, R_S = 0)$. Therefore in the following, to recover $P(V)$, we attempt to express $P(V)$ in terms of the set of observed probabilities $\{P(V_o, S, R_{V_m \setminus S}, R_S = 0) : S \subseteq V_m\}$.

**Example 4** *In Fig. 1(a), instead of the manifest distribution given in Example 1, we work with the set of observed distributions $\{P(X, Y, R_X = 0, R_Y = 0), P(X, R_X = 0, R_Y), P(Y, R_X, R_Y = 0), P(R_X, R_Y)\}$.*

### 4.1 Utility lemmas

The basic idea is to express $P(V)$ and each $P(V_o, S, R_{V_m \setminus S}, R_S = 0)$ in a "canonical" form of factorization of Bayesian networks in terms of c-components [15].

Next we introduce some useful concepts mostly following the notation in [15]. Let $G$ be a Bayesian network structure over $O \cup L$ where $O = \{O_1, \ldots, O_n\}$ is the set of observed variables and $L = \{L_1, \ldots, L_{n'}\}$ is the set of unobserved latent variables. We will often use the notation $G(O, L)$ when we want to make it clear which set of variables in $G$ are latent. For example an m-graph may be denoted by $G(V \cup R, L)$. The observed probability distribution $P(O)$ can be expressed as:

$$P(o) = \sum_l \prod_{\{i|O_i \in O\}} P(o_i|pa_{O_i}) \prod_{\{i|L_i \in L\}} P(l_i|pa_{L_i}), \tag{6}$$

where the summation ranges over all the $L$ variables. For any set $S \subseteq O$, define the quantity $Q[S]$ to denote the following function

$$Q[S](o) = \sum_l \prod_{\{i|O_i \in S\}} P(o_i|pa_{O_i}) \prod_{\{i|L_i \in L\}} P(l_i|pa_{L_i}). \tag{7}$$

In particular, we have $Q[O] = P(o)$. $Q[S]$ is a function of some subset of variables in $O$. For convenience, we will often write $Q[S](o)$ as $Q[S]$.

The set of observed variables $O$ can be partitioned into *c-components* by assigning two variables $O_i$ and $O_j$ to the same c-component if and only if $O_i$ has an unobserved parent $L_i$ and $O_j$ has an unobserved parent $L_j$ such that: either $L_i = L_j$ or there exists a path between $L_i$ and $L_j$ in $G$ such that (i)every internal node of the path is in $L$, *or* (ii) every node in $O$ on the path is head-to-head ($\to O_i \leftarrow$). Note that if an observable variable has no latent parent, then it is a c-component by itself. The importance of the c-components partition lies in that if $O$ is partitioned into c-components $S_1, \ldots, S_k$ then each $Q[S_i]$, called a *c-factor*, is computable in terms of $P(O)$. The following result is from [16, 15]:

**Lemma 1** *Given a DAG $G(O, L)$, assuming that $O$ is partitioned into c-components $S_1, \ldots, S_k$, we have*

*(i) $P(O)$ is decomposed into*

$$P(o) = \prod_i Q[S_i].$$

*(ii) Let a topological order over $O$ be $O_1 < \ldots < O_n$, and let $O^{\leq i} = \{O_1, \ldots, O_i\}$ be the set of variables ordered before $O_i$ (including $O_i$), for $i = 1, \ldots, n$, and $O^{\leq 0} = \emptyset$. Then each $Q[S_j]$, $j = 1, \ldots, k$, is computable from $P(O)$ and is given by*

$$Q[S_j] = \prod_{\{i|O_i \in S_j\}} P(o_i|o^{\leq i-1}). \tag{8}$$

Based on Lemma 1, we have that if $P(v)$ is decomposed into product of c-factors $Q[S_i]$, then $P(v)$ is recovered if each $Q[S_i]$ is recovered.

**Example 5** *In Fig. 1(b), considering the "normal" Bayesian network over $V \cup L$ (the part of the model without $R$ variables), we have*

$$P(x, y, z) = P(x)[\sum_{l_1} P(y|x, l_1) P(z|l_1) P(l_1)] = P(x) Q[\{Y, Z\}]. \tag{9}$$

*Therefore $P(v)$ could be recovered if both $P(x)$ and $Q[\{Y, Z\}]$ are recovered.*

We can also express the given observed distribution $P(V_o, S, R_{V_m \setminus S}, R_S = 0)$ for $S \subseteq V_m$ in its "canonical" form of c-factor factorization based on Lemma 1. Assume $V \cup R$ in $G(V \cup R, L)$ is partitioned into c-components $B_1, \ldots, B_k$, then

$$P(v_o, v_m, r) = \prod_i Q[B_i], \tag{10}$$

and

$$P(v_o, s, r_{V_m \setminus S}, R_S = 0) = \sum_{v_m \setminus s} \prod_i Q[B_i] \Big|_{R_S = 0}. \tag{11}$$

To further utilize Lemma 1 we will consider variables in $V_m \setminus S$ as latent variables, and assume that $V_o \cup S \cup R$ in $G(V_o \cup S \cup R, L \cup (V_m \setminus S))$ is partitioned into c-components $C_1, \ldots, C_m$. Then

$$P(v_o, s, r_{V_m \setminus S}, R_S = 0) = \prod_i Q[C_i]\big|_{R_S=0} \qquad (12)$$

**Example 6** *In Fig. 1(b), we have*[3]

$$P(z, r_X, r_Y, R_Z = 0) = \sum_{x,y} Q[X, R_Z = 0]Q[Y, Z]P(r_X|y)P(r_Y|z) \qquad (13)$$

$$= P(r_Y|z) \sum_{x,y} Q[X, R_Z = 0]Q[Y, Z]P(r_X|y) \qquad (14)$$

In lieu of Lemma 1, we ask a similar question: given the expression in Eq. (27), is a factor $Q[C_i]$ computable in terms of given $P(v_o, s, r_{V_m \setminus S}, R_S = 0)$? The main difference with the situation in Lemma 1 is that variables in $R_S$ are assuming a fixed value.

Next we extend Lemma 1 to the situation that we are not given $P(O)$ but $P(O \setminus S, S = 0)$ for some $S \subset O$. For any set $C$, let $An(C)$ denote the union of $C$ and the set of ancestors of the variables in $C$.

**Lemma 2** *Given a DAG $G(O, L)$, assuming that $O$ is partitioned into c-components $S_1, \ldots, S_k$, we have, for any $S \subseteq O$,*

*(i)*

$$P(O \setminus S, S = 0) = \prod_i Q[S_i]\big|_{S=0}. \qquad (15)$$

*(ii) If $S_j \cap An(S) = \emptyset$, that is, $S_j$ contains no ancestors of $S$, then $Q[S_j]\big|_{S=0}$ is computable from $P(O \setminus S, S = 0)$. In this case, letting a topological order over $O$ be $O_1 < \ldots < O_n$ such that non-acestors of $S$ is ordered after ancestors of $S$, i.e., $An(S) < O \setminus An(S)$, then $Q[S_j]\big|_{S=0}$ is given by*

$$Q[S_j]\big|_{S=0} = \prod_{\{i|O_i \in S_j\}} P(o_i|o^{\leq i-1})\big|_{S=0}. \qquad (16)$$

*Proof:* By Lemma 1, (i) holds and each $Q[S_j]\big|_{S=0}$ can be expressed by Eq. (16). If $S_j$ contains no ancestors of $S$, then all variables in $S_j$ are ordered after $S$. As a consequence $S \subseteq O^{\leq i-1}$ and therefore each term $P(o_i|o^{\leq i-1})\big|_{S=0}$ in Eq. (16) is computable from $P(O \setminus S, S = 0)$. □

**Example 7** *Consider the m-graph in Fig. 1(a). Eq. (27) becomes, for $S = \{Y\}$,*

$$P(y, r_X, R_Y = 0) = P(r_X|R_Y = 0, y)P(y)[\sum_x P(R_Y = 0|x)P(x)] \qquad (17)$$

$R_X$, $Y$, and $R_Y$ each forms a c-component individually in $G(\{Y, R_X, R_Y\}, \{X\})$. By Lemma 2, c-factors $P(r_X|R_Y = 0, y)$ and $P(y)$ are computable from $P(y, r_X, R_Y = 0)$ because neither of $R_X$ or $Y$ is an ancestor of $R_Y$. We also obtain that $[\sum_x P(R_Y = 0|x)P(x)]$ is recoverable by virtue of both $P(r_X|R_Y = 0, y)$ and $P(y)$ being recoverable.

Now for $S = \{X\}$ with $Y$ considered a latent variable,

$$P(x, r_Y, R_X = 0) = P(r_Y|x)P(x)[\sum_y P(R_X = 0|r_Y, y)P(y)], \qquad (18)$$

*none of the three c-factors is computable from $P(x, r_Y, R_X = 0)$ because $R_Y$, $X$, and $R_X$ are all ancestors of $R_X$.*

For $S = \emptyset$ with both $X$ and $Y$ considered latent variables, we have

$$P(r_X, r_Y) = [\sum_x P(r_Y|x)P(x)][\sum_y P(r_X|r_Y, y)P(y)] = Q[R_Y]Q[R_X], \qquad (19)$$

*and both $Q[R_Y]$ and $Q[R_X]$ are computable from $P(R_X, R_Y)$ based on Lemma 1 (or 2).*

---

[3] For convenience, we use $Q[Y, Z]$ to denote $Q[\{Y, Z\}]$ and $Q[X, R_Z = 0]$ to denote $Q[\{X, R_Z\}]\big|_{R_Z=0}$.

**Algorithm** $REJ$

1. Let the c-components of $G_V$ be $A_1, \ldots, A_k$.
2. For every $Q[A_i]$: call REQ($G, P, Q[A_i]$).
3. $P(V)$ is recoverable as $P(v) = \prod_i Q[A_i]$ if every $Q[A_i]$ is recoverable.

Fuction REQ($G', P', Q[C]$)
OUTPUT: Expression for $Q[C]$ or FAIL

1. Assume that $Q[C]$ is a function over $W$. Let $S = W \cap V^m$, $O = V_o \cup S \cup R$.
2. IF $C$ forms a c-component in $G'(O, L \cup (V_m \setminus S))$ and $C \cap An(R_S) = \emptyset$, THEN RETURN $Q[C]$ recoverable as given in Lemma 2.
3. Let $T = (An(C) \cup An(R_S)) \cap O$ and $D = O \setminus T$. IF $D \neq \emptyset$, THEN let $G''$ be the graph resulting from removing $D$ from $G$ and RETURN REQ($G'', \sum_D P', Q[C]$).
4. (a) For each c-component $C_i$ of $G'(O, L \cup (V_m \setminus S))$ such as $C_i \cap An(R_S) = \emptyset$: $Q[C_i]$ is recovered by Lemma 2. Let $G''$ be the graph resulting from removing $C_i$ from $G'$ and RETURN REQ($G'', P'/Q[C_i], Q[C]$).
   (b) For each c-component $C_i$ of $G'(O, L \cup (V_m \setminus S))$ that does not contain $C$ and $Pa(C_i) \cap V^m \neq S$:
   Assume $Q[C_i] = \sum_{v_m \setminus s} \prod_j Q[B_j]$ where each $B_j$ is a c-component of $G(V \cup R, L)$.
   IF $Q[C_i]$ is recovered by REQ($G, P, Q[C_i]$) or every $Q[B_j]$ is recovered by REQ($G, P, Q[B_j]$), THEN let $G''$ be the graph resulting from removing $C_i$ from $G'$ and RETURN REQ($G'', P'/Q[C_i], Q[C]$).
5. RETURN FAIL

Figure 2: Algorithm for recovering $P(V)$.

## 4.2 Recoverability algorithm

Equipped with Lemmas 1 and 2, we are now ready to develop a systematic algorithm for recovering the joint $P(V)$. The basic idea is to first decompose $P(v)$ into product of c-factors $Q[S_i]$, and then attempt to recover each $Q[S_i]$ by applying Lemma 2 to observed probabilities $P(V_o, S, R_{V_m \setminus S}, R_S = 0)$.

**Example 8** *In the m-graph in Fig. 1(a), $P(x, y) = P(x)P(y)$ is recovered if both $P(x)$ and $P(y)$ are recovered. $P(y)$ can be recovered from $P(y, r_X, R_Y = 0)$ as shown in Example 7. However $P(x)$ is not computable from $P(x, r_Y, R_X = 0)$ (see Eq. (18)) because $X$ is an ancestor of $R_X$. On the other hand $Q[R_X]$ has been shown to be computable from $P(R_X, R_Y)$ in Example 7. We rewrite Eq. (18) as:*

$$\frac{P(x, r_Y, R_X = 0)}{Q[R_X]\big|_{R_X=0}} = P(r_Y|x)P(x). \quad (20)$$

*Now $P(x)$ is computable from the recoverable quantity on the left-hand-side of the equation as*

$$P(x) = \sum_{r_Y} \frac{P(x, r_Y, R_X = 0)}{Q[R_X]\big|_{R_X=0}}. \quad (21)$$

*Intuitively, $P(r_Y|x)$ and $P(x)$ are c-factors of the subgraph over $\{R_Y, X, Y\}$ formed by removing the variable $R_X$ from the original m-graph, and both are recoverable by Lemma 1 (or 2).*

For any set $C$, let $G_C$ denote the subgraph of $G$ composed only of variables in $C$. We propose a systematic algorithm REJ for recovering $P(V)$ presented in Fig. 2. REJ works by first decomposing $P(V)$ into product of c-factors $Q[S_i]$, and then attempting to recover each $Q[S_i]$ using a subroutine REQ. REQ works by utilizing Lemma 2 and systematically reducing the problem to simpler one in subgraphs.

**Theorem 1** *Algorithm REJ is sound*

The proof of Theorem 1 is given in the Supplementary Material.

**Example 9** *In the m-graph in Fig. 1(b), we want to recover $P(x,y,z) = P(x)Q[Y,Z]$. First we attempt to recover $Q[Y,Z]$ from*

$$P(x,y,z,R_{X,Y,Z}=0) = Q[X,R_Z=0]Q[Y,Z]P(R_X=0|y)P(R_Y=0|z). \quad (22)$$

*Step 4 of REQ calls for recovering $P(R_Y=0|z)$ from $P(z,r_X,r_Y,R_Z=0)$ given in Eq. (14). We have that $P(r_Y|z)$ is recoverable from the base case Step 2 by Lemma 2. REQ Step 4 also calls for recovering $P(R_X=0|y)$ from*

$$P(y,r_X,r_Z,R_Y=0) = P(r_X|y) \sum_{x,z} Q[X,R_Z]Q[Y,Z]P(R_Y=0|z). \quad (23)$$

*Again we have that $P(r_X|y)$ is recoverable from the base case Step 2 by Lemma 2. Then Step 4 reduces the problem to recovering $Q[Y,Z]$ from*

$$\frac{P(x,y,z,R_{X,Y,Z}=0)}{P(R_Y=0|z)P(R_X=0|y)} = Q[X,R_Z=0]Q[Y,Z]. \quad (24)$$

*We obtain that $Q[Y,Z]$ is recoverable from the base case Step 2 by Lemma 2.*

*Next we use REQ to recover $P(x)$ from*

$$P(x,r_Y,r_Z,R_X=0) = Q[X,R_Z] \sum_{y,z} Q[Y,Z]P(R_X=0|y)P(r_Y|z). \quad (25)$$

*Step 3 reduces the problem to recovering $P(x)$ from, by summing out $R_Z$ and $R_Y$,*

$$P(x,R_X=0) = P(x) \sum_{y,z} Q[Y,Z]P(R_X=0|y). \quad (26)$$

*We have shown both $Q[Y,Z]$ and $P(R_X=0|y)$ are recoverable, and therefore $P(x)$ is recoverable. In conclusion $P(x,y,z)$ is recoverable.*

It is natural to ask whether the algorithm REJ presented in Fig. 2 is complete, that is whether the output of FAIL corresponds to that $P(V)$ is not recoverable. We are not able to answer this difficult question at this point. We find the algorithm promising in that it has pinned down situations in which recoverability seems not possible. We consider the result a significant advance over the existing sufficient conditions in the literature [1, 2].

## 5 Conclusion

It is of theoretical interest and importance in practice to determine in principle whether a probabilistic query is estimable from missing data or not when the data are not MAR. In this paper we present an algorithm for systematically determining the recoverability of the joint distribution from observed data with missing values given an m-graph with latent variables. The result is significantly more general than the sufficient conditions in [1, 2]. Compared to the result in [11] for Markov models, we allow latent variables in the model, and treat the problem in a purely probabilistic framework without appealing to causality theory. Our algorithm is of course applicable to Markov models, for which the algorithm could be simplified. We have also developed new simple sufficient conditions that could be used to quickly recover the joint in Markov models. These results on recovering the joint distribution in Markov models are presented in the Supplementary Material. Future work includes developing algorithms for recovering arbitrary probabilistic queries such as $P(x|y)$, and for recovering causal queries such as $P(y|do(x))$. It is also an interesting research direction how to actually estimate distribution parameters from finite amount of data if the joint is determined to be recoverable [17].

# 6 Supplementary Material

## 6.1 Proof of Theorem 1

**Theorem 2** *Algorithm REJ is sound*

*Proof:* Based on Lemma 1, REJ is sound if REQ is sound. Next we show the soundness of REQ.

Step 1 specifies the smallest $S$ such that $Q[C]$ could potentially form a c-factor in $G(O, L \cup (V_m \setminus S))$. The algorithm then attempts to recover $Q[C]$ from observed $P(V_o, S, R_{V_m \setminus S}, R_S = 0)$.

Step 2 is the base case which is sound based on Lemma 2.

In Step 3 summing out $D$ from both sides of the following (Eq. (12) in the main paper)

$$P(v_o, s, r_{V_m \setminus S}, R_S = 0) = \prod_i Q[C_i] \big|_{R_S=0} \tag{27}$$

is graphically equivalent to removing variables in $D$ based on the chain rule of Bayesian networks since all $D$ variables could be ordered after $T$ variables.

In Step 4 we attempt to recover another c-factor $Q[C_i]$, either from $P(V_o, S, R_{V_m \setminus S}, R_S = 0)$ by Lemma 2 in 4(a) or from other observed distributions by calling REQ in 4(b). If $Q[C_i]$ is recovered, then given

$$P(v_o, s, r_{V_m \setminus S}, R_S = 0) = Q[C_i] \prod_{j \neq i} Q[C_j] \big|_{R_S=0}, \tag{28}$$

we obtain

$$\frac{P(v_o, s, r_{V_m \setminus S}, R_S = 0)}{Q[C_i]} = \prod_{j \neq i} Q[C_j] \big|_{R_S=0}. \tag{29}$$

Now the problem of recovering $Q[C]$ is reduced to a problem of recovering $Q[C]$ in the subgraph resulting from removing $C_i$ from $G$ with associated distribution $\frac{P(v_o, s, r_{V_m \setminus S}, R_S=0)}{Q[C_i]}$.

We fail to recover $Q[C]$ if $Q[C]$ cannot be recovered by Lemma 2 and the problem cannot be reduced to a smaller one by Steps 3 or 4. □

## 6.2 Recoverability in Markov models

If the m-graph does not contain latent variables ($L = \emptyset$), the model is called a Markov model. In this situation, we have

$$P(v) = \prod_i P(v_i|pa_i), \tag{30}$$

and

$$P(v_o, v_m, R = 0) = P(v)P(R = 0|v) = \prod_i P(v_i|pa_i) \prod_j P(R_{V_j} = 0|pa_{r_{V_j}}) \big|_{R=0} \tag{31}$$

where $Pa_i$ and $Pa_{r_{V_j}}$ represent the parents of $V_i$ and $R_{V_j}$ in $G$ respectively. We have that $P(V)$ is recoverable if every factor $P(v_i|pa_i)$ is recoverable. Alternatively, from Eq. (31) we have that $P(V)$ is recoverable if every factor $P(R_{V_j} = 0|pa_{r_{V_j}})$ is recoverable. The algorithm REJ will recover $P(v)$ by attempting to recover every $P(v_i|pa_i)$. We observe that it is often the case that to recover all $P(v_i|pa_i)$ the algorithm REQ normally will need to recover many $P(R_{V_j} = 0|pa_{r_{V_j}})$ for $R_{V_j}$ being a descendant of some $V_i$ (the opposite is not true). Therefore for the purpose of recovering $P(v)$ it is often more efficient to recover all $P(R_{V_j} = 0|pa_{r_{V_j}})$ instead. We observe that the MID algorithm in [11] appears to be recovering $P(v)$ by recovering every $P(R_{V_j} = 0|pa_{r_{V_j}})$.

There exist simple sufficient conditions by which we can quickly recover $P(R_{V_j} = 0|pa_{r_{V_j}})$. For example a necessary condition for $P(R_{V_j} = 0|pa_{r_{V_j}})$ being recoverable is that $V_j$ is not a parent of $R_{V_j}$ [1]. We summarize several sufficient conditions in the following proposition.

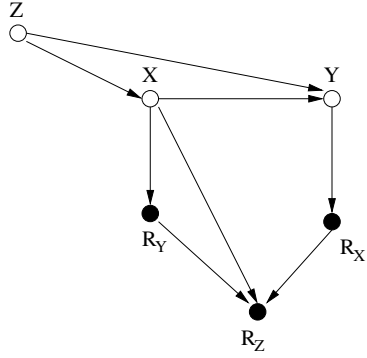

Figure 3: $P(X, Y, Z)$ is recoverable.

**Algorithm** REJ-M

1. For every $V_j \in V_m$:
   Recover $P(R_{V_j} = 0|pa_{r_{V_j}})\big|_{R=0}$ by Proposition 2 if applicable,
   otherwise call REQ$(G, P, P(R_{V_j} = 0|pa_{r_{V_j}}))$.
2. $P(V)$ is recoverable if every $P(R_{V_j} = 0|pa_{r_{V_j}})$ is recoverable.

Figure 4: Algorithm for recovering $P(V)$ in Markov models.

**Proposition 2** $P(R_{V_j}|pa^o_{r_{V_j}}, pa^m_{r_{V_j}}, pa^r_{r_{V_j}})$, where $Pa^o_{r_{V_j}}$, $Pa^m_{r_{V_j}}$, and $Pa^r_{r_{V_j}}$ are the parents of $R_{V_j}$ in G that are $V^o$ variables, $V^m$ variables, and R variables respectively, is not recoverable if $V_j$ is a parent of $R_{V_j}$; otherwise, $P(R_{V_j}|pa^o_{r_{V_j}}, pa^m_{r_{V_j}}, pa^r_{r_{V_j}})\big|_{R=0}$ is recoverable if one of the following holds:

1. $Pa^m_{r_{V_j}} = \emptyset$.
2. $Pa^m_{r_{V_j}}$ is a subset of the set of $V_m$ variables corresponding to $Pa^r_{r_{V_j}}$.
3. $R_{V_j}$ has no child.
4. None of $R_{Pa^m_{r_{V_j}}}$ is a descendant of $R_{V_j}$.

*Proof:* Conditions 1 and 2 are straightforward and used extensively in [1, 9, 2].

Conditions 3 and 4: $P(R_{V_j}|pa^o_{r_{V_j}}, pa^m_{r_{V_j}}, pa^r_{r_{V_j}})$ is a c-factor in $P(V_o, S, R_{V_m \setminus S}, R_S = 0)$ for $S = Pa^m_{r_{V_j}}$. Then $P(R_{V_j}|pa^o_{r_{V_j}}, pa^m_{r_{V_j}}, pa^r_{r_{V_j}})$ is recoverable by Lemma 2 since $R_{V_j}$ is not an ancestor of $R_S$. □

Based on the above conditions 3 and 4, we present the following sufficient condition for recovering $P(V)$.

**Theorem 3** *In a Markov model $P(V)$ is recoverable if no variable X is a parent of its corresponding $R_X$, and for each $R_X$, either it has no child, or none of the R variables correponding to its $V_m$ parents are descendants of $R_X$.*

**Example 10** $P(X, Y, Z)$ *is recoverable in Fig. 3 by Theorem 3.*

In general we propose a systematic algorithm REJ-M for recovering $P(V)$ in Markov models presented in Fig. 4.

**Example 11** *In the m-graph in Fig. 5(a), we attempt to recover $P(A, B, C, D)$ by recovering $P(R_A = 0|d, b, R_B = 0)$ and $P(R_B = 0|d, a)$. First $P(R_A = 0|d, b, R_B = 0)$ is easily re-*

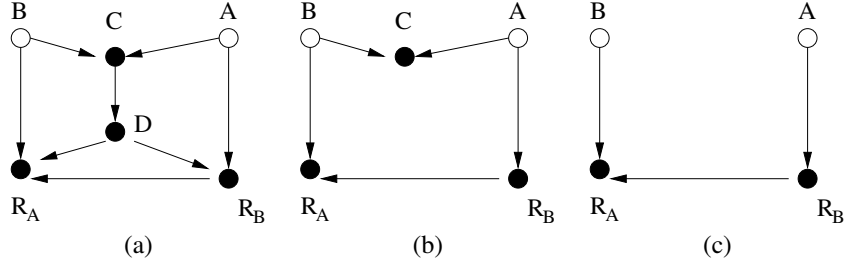

Figure 5: $P(A, B, C, D)$ is recoverable.

*covered by condition 2 or 3 in Proposition 2. Next we call REQ($G, P, P(R_B = 0|d, a)$), which attempts to recover $P(R_B = 0|d, a)$ from ($S = \{A\}$)*

$$P(c, d, a, r_B, R_A = 0) = P(a)P(d|c)P(r_B|d, a)[\sum_b P(b)P(c|a, b)P(R_A = 0|d, b, r_B)]. \quad (32)$$

*In Step 4 of REQ we attempt to recover $P(d|c)$ from*

$$P(c, d, r_A, r_B) = P(d|c) \sum_{a,b} P(a)P(r_B|d, a)P(b)P(c|a, b)P(r_A|d, b, r_B), \quad (33)$$

*which says $P(d|c)$ is recoverable by Lemma 2. The problem is reduced to recovering $P(R_B = 0|d, a)$ in Fig. 5(b) from*

$$\frac{P(c, d, a, r_B, R_A = 0)}{P(d|c)} = P(a)P(r_B|d, a)[\sum_b P(b)P(c|a, b)P(R_A = 0|d, b, r_B)]. \quad (34)$$

*$C$ is not an ancestor of $R_B$ or $R_A$ in Fig. 5(b), and Step 3 of REQ reduces the problem to recovering $P(R_B = 0|d, a)$ in Fig. 5(c) from*

$$\sum_c \frac{P(c, d, a, r_B, R_A = 0)}{P(d|c)} = P(a)P(r_B|d, a)[\sum_b P(b)P(R_A = 0|d, b, r_B)]. \quad (35)$$

*Fig. 5(c) is the same as Fig. 1(a) for which $P(r_B|d, a)$ can be recoverable as shown in Example 7. In fact, $Q[R_A = 0] = \sum_b P(b)P(R_A = 0|d, b, r_B)$ can be recovered from $P(c, d, r_A, r_B)$ from the following*

$$\sum_c \frac{P(c, d, r_A, r_B)}{P(d|c)} = [\sum_a P(a)P(r_B|d, a)][\sum_b P(b)P(R_A|d, b, r_B)], \quad (36)$$

*from which $Q[R_A]$ is recoverable by Lemma 2 (or 1). Finally the problem is reduced to recovering $P(R_B = 0|d, a)$ from the following*

$$\frac{1}{Q[R_A = 0]} \sum_c \frac{P(c, d, a, r_B, R_A = 0)}{P(d|c)} = P(a)P(r_B|d, a), \quad (37)$$

*from which it is clear $P(R_B = 0|d, a)$ is recoverable by Lemma 2.*

This example is used to demonstrate the MID algorithm in [11]. We suspect our REJ-M algorithm works in a somewhat similar way as MID, but we use a pure probabilistic framework without appealing to causality techniques.